\documentclass[letterpaper,10pt,conference]{ieeeconf}
\IEEEoverridecommandlockouts
\usepackage{amsmath,amssymb}
\usepackage{graphicx}
\usepackage[table]{xcolor}
\usepackage{booktabs}
\usepackage{cite}
\usepackage{url}
\usepackage{microtype}
\usepackage{placeins}
\graphicspath{{figures/}}
\newcommand{\zz}[1]{\,\mathrm{#1}}
\definecolor{c0}{rgb}{0.25,0.25,0.25}
\newcommand{\tw}{0.82\columnwidth}

\title{\LARGE \bf Task-Oriented Co-Design and Optimization of Geared Actuators for Robotic Applications}
\author{Xuanyu Huang, Jianqiang Dong, and Hang Zhao%
\thanks{This work was supported by the National Natural Science Foundation of China under Grant 52407066.}%
\thanks{The authors are with the Hong Kong University of Science and Technology (Guangzhou), Guangzhou, China
(e-mail: xhuang313@connect.hkust-gz.edu.cn; jdong921@connect.hkust-gz.edu.cn; hangzhao@hkust-gz.edu.cn). Hang Zhao is the corresponding author.}}

\begin{document}
\maketitle
\thispagestyle{empty}
\pagestyle{empty}
\begin{abstract}
Different tasks performed by legged robots impose distinct torque and speed requirements on actuators.
Existing robotic actuators are generally optimized at the component level for metrics such as torque or power density, without explicit task guidance.
System-level optimization across components such as motors, gearboxes, and sensors is challenging because of the high computational cost and coupling among mechanical, electrical, and electromagnetic behaviors.
Consequently, improvements in individual components may not translate into better robot performance in a specific task.
To this end, we present a systematic optimization framework for task-oriented co-design of actuator hardware and control.
First, surrogate models are employed to accelerate motor evaluation and support global exploration of the coupled design space.
Then, a hierarchical mixed-variable optimization strategy is adopted, combining discrete enumeration with continuous search over dimensions and real-valued indices.
These indices are rounded to select admissible values for the remaining discrete choices before each evaluation.
Within this search, rated output torque density and task performance are jointly optimized, with B\'ezier-parameterized joint torque profiles determined for each hardware candidate.
Finally, the effectiveness of the proposed framework is validated through actuator fabrication and experiments on a two-degree-of-freedom jumping leg.
Based on its measured mass, the fabricated prototype achieves a nominal rated output torque density of $35.7\zz{N\,m/kg}$, approximately $60\%$ higher than that of a widely used commercial geared joint actuator, while being $18.6\%$ lighter.
Under matched bench conditions, it achieves $12.0\%$ greater jump height at twice-rated torque.
Together, these results demonstrate a systematic route from task requirements to actuator design and control.
\end{abstract}


\section{Introduction}
Advances in permanent-magnet motors, power electronics, batteries, and control strategies have made compact electric joint drives central to legged and humanoid robots \cite{r01,r02}. Early electric humanoids generally relied on compact high-reduction servo joints and rigid position control \cite{r03,r04,r05,r06,r07,r08}. Series-elastic and low-reduction drives later emphasized force regulation, impact tolerance, transparency, and bandwidth \cite{r09,r10,r11}, while the Mini Cheetah demonstrated highly integrated modular drives for dynamic motion \cite{r12}. These architectures entail distinct trade-offs: high reduction increases compact output torque but also reflected inertia and transmission losses \cite{r13}; series elasticity adds mechanical and control complexity; and low reduction improves backdrivability but demands higher motor torque density \cite{r14,r15}. Their suitability therefore depends on the robot scale, package, and intended motion rather than on a single component metric. Among these architectures, this work focuses on compact actuators based on planetary gearboxes, which are widely adopted for their practical balance of cost, control simplicity, and torque density \cite{r12,r02,r15}.

\begin{figure}[!t]
  \centering\includegraphics[width=\columnwidth]{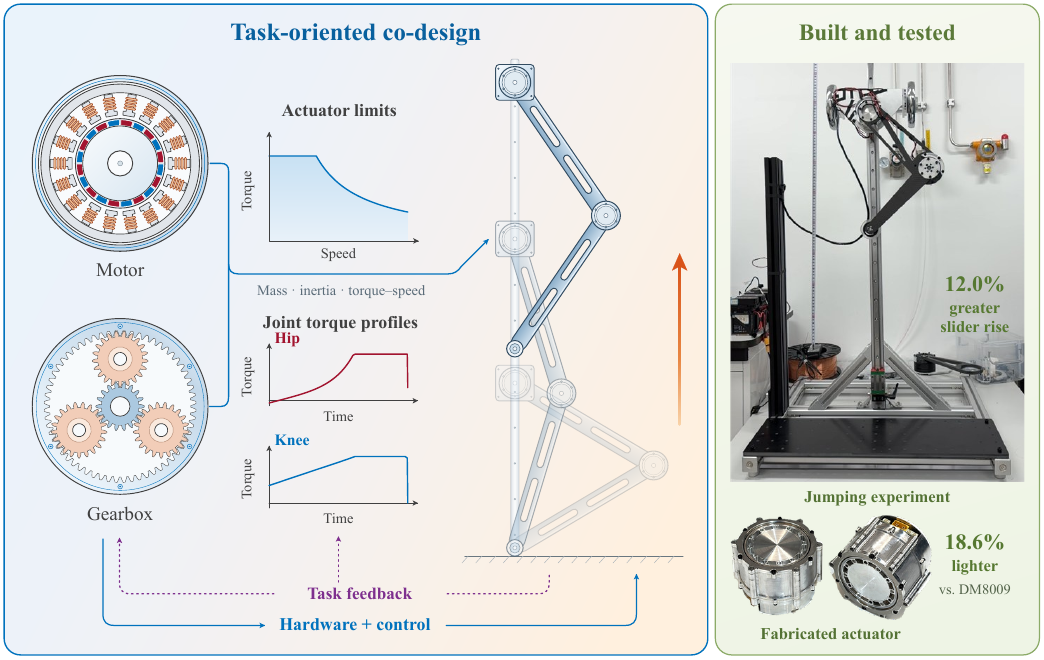}
  \caption{Overview of the task-oriented motor--gearbox and control co-design framework and its experimental validation.}
  \label{l001}
\end{figure}

Robots using such actuators have progressed from steady locomotion to increasingly dynamic motions such as running and jumping. These motions do not merely scale a common actuator requirement: even on the same robot, they occupy different torque--speed regions and emphasize different combinations of continuous torque, peak torque, speed, efficiency, mass, and reflected inertia. Component ratings alone therefore cannot determine whether an actuator is well matched to a particular motion. Task-aware design and robot co-design studies recognize this dependence \cite{r16,r17,r18,r19}. Task-oriented co-design is therefore needed to match actuator characteristics to task-specific demands and improve robot-level performance across diverse operating scenarios.

Closing this task-to-actuator loop requires motor, gearbox, and robot dynamics to be considered together. Motor geometry, transmission topology, reduction ratio, reflected inertia, mass, and control limits jointly define the feasible joint-level operating envelope \cite{r13,r14}. Drive-train optimization has jointly selected motors and transmissions for system-level performance \cite{r20}; computational design has extended co-design to leg geometry and actuator sizing \cite{r21,r17}; and dynamic co-design has coupled robot morphology, actuation, and motion \cite{r22,r18}. However, integrating detailed motor and gearbox design, task dynamics, and control within one optimization remains difficult: the models span electromagnetic, electrical, mechanical, and control domains; the design variables combine discrete and continuous choices; and each candidate is costly to evaluate. Many existing co-design formulations therefore fix motor characteristics or use simplified scaling laws. Such assumptions reduce computational cost but leave detailed motor electromagnetic and gearbox design variables outside the system-level optimization. To address this limitation, the proposed framework integrates these variables with task dynamics and candidate-specific control; surrogate motor evaluation \cite{r23,r24} and hierarchical mixed-variable optimization keep the coupled search tractable.

Figure~\ref{l001} summarizes the proposed task-oriented co-design and validation loop, coupling motor and gearbox design with joint torque control and jumping-task feedback.

This paper makes three contributions. First, it formulates task-oriented actuator co-design by linking motor geometry and winding, planetary-gearbox design, actuator mass and inertia, and the dynamics of a reduced-order two-degree-of-freedom jumping leg. Second, it develops a surrogate-assisted hierarchical mixed-variable optimization strategy that enumerates major discrete cases and performs continuous search over bounded motor and gearbox variables, with real-valued indices rounded to admissible values for the remaining discrete choices. Third, it couples hardware search with candidate-specific control-policy optimization through B\'ezier-parameterized hip and knee torque profiles. The selected design is then fabricated and experimentally validated against a commercial geared actuator.

\section{Task Motivation and Jumping-Leg Model}
\subsection{Task-Dependent Requirements in Humanoid Motion}

\begin{figure}[!t]
  \centering
  \includegraphics[width=\columnwidth]{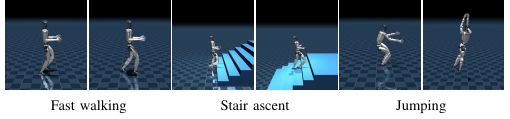}
  \caption{Representative MuJoCo frames for Unitree G1 fast walking, stair ascent, and jumping.}
  \label{l002}
  \vspace{0.35em}
  \includegraphics[width=\columnwidth]{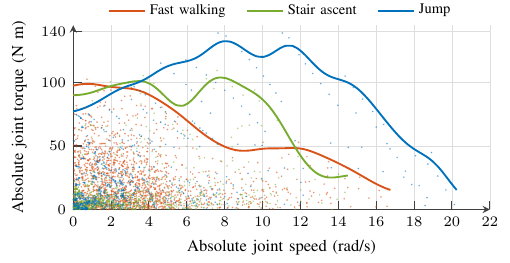}
  \caption{Pooled G1 knee and hip-pitch torque--speed operating points and smoothed moving-window requirement envelopes for fast walking, jumping, and stair ascent. Envelopes use all recorded samples; scatter points are subsampled for visibility.}
  \label{l003}
\end{figure}

The G1 data used here comprise fast walking, a dynamic jumping takeoff, and a six-step stair-ascent rollout, all simulated with the Unitree G1 model in MuJoCo \cite{r25}. For each motion, the bilateral knee and hip-pitch torque--speed samples are mapped to absolute magnitudes and pooled into one task-level point set. Requirement envelopes are computed as moving-window maxima over $1.0\zz{rad/s}$ speed neighborhoods and smoothed over a $0.75\zz{rad/s}$ bandwidth for visualization.

Fig.~\ref{l003} shows that jumping occupies the highest torque--speed region. Across the pooled knee and hip-pitch samples, the peak torque and speed are $103\zz{N\,m}$ and $16.6\zz{rad/s}$ for fast walking, $139\zz{N\,m}$ and $20.1\zz{rad/s}$ for jumping, and $109\zz{N\,m}$ and $14.3\zz{rad/s}$ for stair ascent. The fast-walking and stair-ascent envelopes cross: fast walking extends to higher speed, whereas stair ascent reaches higher torque. These data show that different dynamic tasks impose distinct torque--speed requirements on their actuators.

\subsection{Reduced-Order Jumping-Leg Task}
\begin{figure}[!t]
  \centering
  \includegraphics[width=\columnwidth]{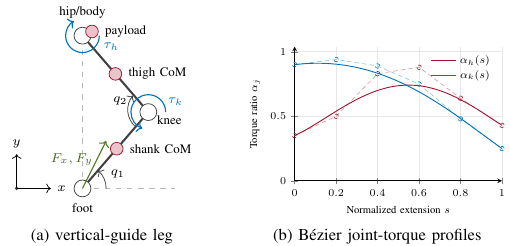}
  \caption{Reduced-order jumping-task representation: (a) the two-DOF vertical-guide leg and (b) representative fifth-order B\'ezier profiles with six control coefficients for each joint.}
  \label{l004}
\end{figure}

The G1 results reveal a key actuator-level demand of jumping: high joint torque
over a broad speed range. To demonstrate the effectiveness of the proposed
co-design framework while keeping computation and hardware validation tractable,
we reduce the whole-robot jumping problem to the two-DOF vertical-guide leg in
Fig.~\ref{l004}(a), yielding a leg-level actuator co-design benchmark
\cite{r17}. The model retains coupled hip--knee push-off, ground
contact, and the resulting joint torque--speed trajectories while removing
whole-body balance and lateral dynamics. The framework is not specific to this
benchmark and can be extended to other robots and tasks by replacing the task
model, objectives, and task-specific constraints within the same co-design loop.

\begin{figure*}[!t]
  \centering
  \includegraphics[width=\textwidth]{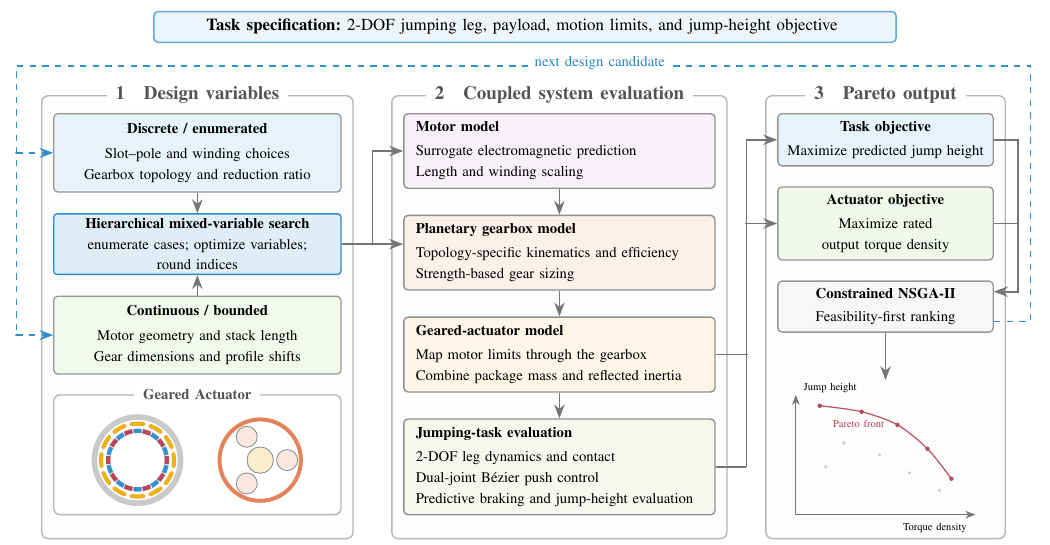}
  \caption{Hierarchical mixed-variable optimization for task-oriented motor--gearbox co-design and constrained Pareto selection.}
  \label{l005}
\end{figure*}

Let $\mathbf{q}=[q_1,q_2]^T$, where $q_1$ is the absolute shank angle and
$q_2$ is the included knee angle. During sticking push-off, the foot is taken
as the origin. For equal shank and thigh lengths, $l_s=l_t$, the vertical guide
imposes
\begin{equation}
\begin{aligned}
x_h(\mathbf{q})
  &=l_s\cos q_1-l_t\cos(q_1-q_2)=0,\\
y_h(\mathbf{q})
  &=l_s\sin q_1-l_t\sin(q_1-q_2).
\end{aligned}
\label{l006}
\end{equation}
With $\mathbf{A}(\mathbf{q})=\partial x_h/\partial\mathbf{q}$ and guide reaction
$\lambda$, let $\boldsymbol{\tau}=[\tau_h,\tau_k]^T$ collect the hip and knee
actuator output torques, with each torque defined positive in the leg-extension
direction. The constrained stance dynamics and actuator-to-generalized torque
mapping are
\begin{equation}
\mathbf{M}(\mathbf{q})\ddot{\mathbf{q}}+\mathbf{h}+\mathbf{G}
=\mathbf{B}\boldsymbol{\tau}+\mathbf{A}(\mathbf{q})^T\lambda,
\quad
\mathbf{B}\boldsymbol{\tau}
=\begin{bmatrix}-\tau_h\\ \tau_h+\tau_k\end{bmatrix}.
\label{l007}
\end{equation}
Motivated by Ding and Park \cite{r17}, we parameterize both joint
torque ratios with separate B\'ezier curves over the common normalized
extension phase $s$, as illustrated in Fig.~\ref{l004}(b):
\begin{equation}
\begin{aligned}
s&=\frac{q_2-q_{2,0}}{q_{2,f}-q_{2,0}},
&j&\in\{h,k\},\\
\tau_j(s)&=\alpha_j(s)\bar{\tau}_j,\\
\alpha_j(s)&=\sum_{i=0}^{5}c_{j,i}\binom{5}{i}(1-s)^{5-i}s^i,
&0&\leq c_{j,i}\leq1.
\end{aligned}
\label{l008}
\end{equation}
Here, $\bar{\tau}_j$ is the available torque at joint $j$ under its
instantaneous actuator limits. The policy contains 12 free coefficients, six
for each joint. For every motor--gearbox candidate, all 12 coefficients are
jointly optimized using bounded L-BFGS-B to maximize $h_{\mathrm{jump}}$ while
the contact constraints remain enforced during the rollout.

To prevent either joint from reaching its travel limit at high speed, an
inertia-aware predictor initiates coordinated braking when
\begin{equation}
\begin{aligned}
d_{\mathrm{stop},j}
  &=\omega_j^{+}t_d
    +\frac{J_{\mathrm{eff},j}(\omega_j^{+})^2}{2\tau_c}
    +\Delta\phi_m,\\
d_{\mathrm{rem},j}&\leq d_{\mathrm{stop},j},
\qquad j\in\{h,k\},
\end{aligned}
\label{l009}
\end{equation}
where $\omega_j^{+}$ is the extension-direction speed, $J_{\mathrm{eff},j}$
includes the reflected rotor inertia, $\tau_c$ is the available braking torque,
$t_d$ is the response-delay allowance, and $\Delta\phi_m$ is a position margin.
Once either axis satisfies Eq.~\eqref{l009}, coordinated reverse
torques brake both joints before the mechanical limit. Throughout push-off and
catch, the actuator torques are bounded by the motor--gearbox-dependent
torque--speed envelopes, while sticking contact requires $F_n\geq0$ and
$|F_t|\leq\mu F_n$. The equations are integrated until release, and the
terminal stance state determines the subsequent vertical motion. The task
metric is
\begin{equation}
\begin{aligned}
h_{\mathrm{jump}}
  &=\max_{t\geq0}\left[y_h(t)-y_h(0)\right]\\
  &\equiv\mathcal{F}(\mathbf{x}_m,\mathbf{x}_g,\mathbf{c}).
\end{aligned}
\label{l010}
\end{equation}
Here, $\mathcal{F}$ denotes the coupled nonlinear response under fixed task
conditions; $\mathbf{x}_m$, $\mathbf{x}_g$, and $\mathbf{c}$ collect the motor,
gearbox, and B\'ezier-policy variables, respectively. The quantity $y_h$ is the
vertical hip position and, under the guide constraint, also the slider position
on the experimental bench; we therefore refer to $h_{\mathrm{jump}}$ as the
slider rise.

\section{Motor--Gearbox Co-Design Method}
Here, motor--gearbox co-design denotes the joint search over motor
electromagnetic design and gearbox topology and design parameters. Each
combined design is evaluated through its resulting actuator mass, reflected
inertia, torque--speed limits, and predicted task-level jumping performance.
With the representative jumping task fixed, this section formulates the
resulting mixed discrete--continuous problem and describes its
surrogate-assisted evaluation and constrained Pareto search.

\begin{table}[!t]
\caption{Motor--Gearbox Design Variables and Constraints}
\label{l011}
\centering
\begingroup
\renewcommand{\arraystretch}{1.10}
\setlength{\tabcolsep}{4pt}
\newcommand{\tc}[1]{\begin{tabular}[t]{@{}l@{}}#1\end{tabular}}
\resizebox{\columnwidth}{!}{%
\begin{tabular}{@{}llll@{}}
\toprule
\tc{Group} & \tc{Symbols} & \tc{Range or\\options} &
\tc{Feasibility\\constraints}\\
\midrule
\tc{Motor:\\slot--pole} & \tc{$(S,P)$} &
\tc{$18\mathrm{S}/\{16,20,22\}\mathrm{P}$;\\
$24\mathrm{S}/\{22,26,28\}\mathrm{P}$;\\
$27\mathrm{S}/30\mathrm{P}$} &
\tc{winding factor:\\$k_{w,1}>0.9$}\\
\rowcolor{c0!5}
\tc{Motor:\\radial} &
\tc{$D_{\mathrm{so}},D_{\mathrm{b}}$;\\$w_t,h_s$} &
\tc{$D_{\mathrm{so}}:60$--$96\zz{mm}$;\\
$D_{\mathrm{b}}:36.1$--$75\zz{mm}$;\\
$w_t:2$--$5\zz{mm}$; $h_s:1$--$19.15\zz{mm}$} &
\tc{valid stator and\\slot geometry}\\
\tc{Motor:\\slot/magnet} &
\tc{$h_{\mathrm{tt}},w_{\mathrm{so}}$;\\
$t_{\mathrm{mag}},\alpha_{\mathrm{mag}}$} &
\tc{$h_{\mathrm{tt}},w_{\mathrm{so}}:0.5$--$2\zz{mm}$;\\
$t_{\mathrm{mag}}:1$--$3\zz{mm}$;\\
$\alpha_{\mathrm{mag}}:120$--$180^\circ$ elec.} &
\tc{positive clearances;\\manufacturable geometry}\\
\rowcolor{c0!5}
\tc{Motor:\\axial/winding} &
\tc{$l_m,d_w,b$} &
\tc{$l_m:1$--$30\zz{mm}$;\\standard wire index;\\symmetry divisors} &
\tc{realizable winding;\\current/voltage limits}\\
\tc{Reducer:\\selection} &
\tc{$\mathcal{T},i_t$} &
\tc{$\mathcal{T}\in\{\mathrm{NGW,NW,3K}\}$;\\
$i_t:1$--$50$ in $0.5$ steps} &
\tc{assembly and ratio\\feasibility}\\
\rowcolor{c0!5}
\tc{Reducer:\\geometry} &
\tc{$\mathbf{z},r_a$;\\$\mathbf{m},\mathbf{x}_p$} &
\tc{topology-dependent integers;\\bounded continuous values} &
\tc{geometry and tooth-root\\strength constraints}\\
\tc{Actuator:\\package} &
\tc{$D_{\mathrm{act}},L_{\mathrm{act}}$} &
\tc{derived} &
\tc{$D_{\mathrm{act}}\leq98\zz{mm}$;\\
$L_{\mathrm{act}}\leq50\zz{mm}$}\\
\rowcolor{c0!5}
\tc{Task:\\contact} &
\tc{$F_n,F_t$} &
\tc{derived} &
\tc{$F_n\geq0$; $|F_t|\leq\mu F_n$;\\valid task integration}\\
\bottomrule
\end{tabular}%
}
\endgroup
\end{table}

\begin{figure}[!t]
  \centering
  \includegraphics[width=\columnwidth]{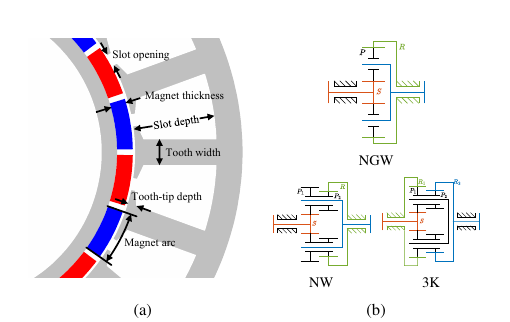}
  \caption{Coupled motor--gearbox design space: (a) motor geometric design variables and (b) planetary gearbox topologies considered, including NGW, NW, and 3K configurations.}
  \label{l012}
\end{figure}

\subsection{Problem Formulation}
Let $\mathbf{x}\in\mathcal{X}_{\mathrm{mix}}$ denote a candidate geared-actuator
design. Its motor, reducer, assembled-actuator, and jumping-task responses are
evaluated by the coupled map $\boldsymbol{\Phi}(\mathbf{x})$ shown in
Fig.~\ref{l005}.

\subsubsection{Objectives}
The task-level objective is the predicted slider rise
$h_{\mathrm{jump}}$, while the actuator-level objective is the rated output
torque density $\rho_T$. Rated torque density is a common actuator-level measure
of mass-specific torque capability in dynamic legged robots
\cite{r10,r11}. These objectives are retained
separately because jump height captures hardware--control interactions that
rated torque density alone cannot represent. Following the minimization
convention used by NSGA-II, the problem is written as
\begin{equation}
\begin{aligned}
\underset{\mathbf{x}\in\mathcal{X}_{\mathrm{mix}}}{\operatorname{minimize}}\quad
\mathbf{f}(\mathbf{x})
  &=\left[-h_{\mathrm{jump}}(\mathbf{x}),-\rho_T(\mathbf{x})\right],\\
\rho_T(\mathbf{x})
  &=\frac{T_{\mathrm{rated,out}}(\mathbf{x})}{m_{\mathrm{act}}(\mathbf{x})}.
\end{aligned}
\label{l013}
\end{equation}
For consistent comparison, $T_{\mathrm{rated,out}}$ is evaluated for all
candidates at the same single-conductor current-density setting,
$J_{\mathrm{rated}}=6\zz{A/mm^2}$, a commonly adopted value for naturally
cooled permanent-magnet machines
\cite{r26,r27}.

\subsubsection{Variables}
The motor and gearbox vectors introduced above are partitioned into discrete
and continuous components in the hardware design vector,
$\mathbf{x}=[\mathbf{x}_{m}^{d},\mathbf{x}_{m}^{c},
\mathbf{x}_{g}^{d},\mathbf{x}_{g}^{c}]$. Motor variables describe the
slot--pole layout, radial and axial geometry, magnets, and winding. Reducer
variables describe the planetary topology, target ratio, tooth numbers, and
geometry.
The combined variables and their associated feasibility checks are summarized in
Table~\ref{l011}; their geometric interpretation is
shown in Fig.~\ref{l012} \cite{r28,r13}.

\begin{figure}[!t]
  \centering
  \includegraphics[width=\columnwidth]{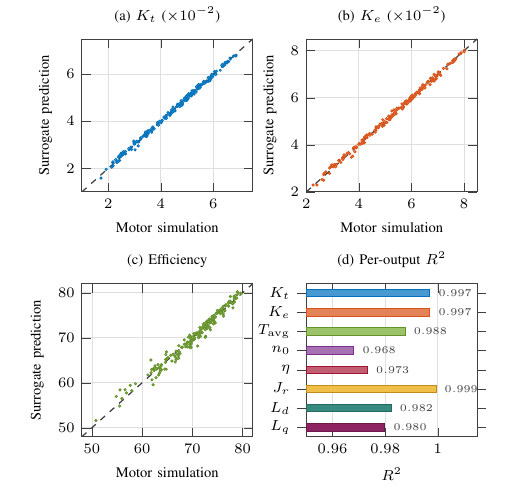}
  \caption{Validation of the 18S/20P motor surrogate using 200 independent cases: parity plots for $K_t$, $K_e$, and efficiency, and $R^2$ values for eight predicted outputs.}
  \label{l014}
\end{figure}

\subsubsection{Constraints}
A candidate is feasible only if its motor geometry and winding are realizable,
its operating points satisfy current and voltage limits, and its reducer meets
assembly, clearance, and tooth-root-strength requirements
\cite{r29}. The assembled actuator must also satisfy the reference
package envelope, and the jumping simulation must maintain admissible contact
and complete without an invalid state. The corresponding checks are summarized
in Table~\ref{l011}.

\subsection{Hierarchical Mixed-Variable Search}
The complete formulation couples more than 20 discrete and continuous
hardware--control variables. We handle this mixed space hierarchically. Major
categorical choices, including slot--pole combinations and reducer topologies,
are first enumerated. Within each enumerated case, the bounded continuous motor
and gearbox variables are searched directly. Remaining discrete choices are
represented by real-valued indices, then clipped, rounded, and decoded into
admissible values before candidate evaluation. This preserves real-valued
resolution for continuous variables while ensuring that every decoded discrete
choice is realizable.

The decoded motor and gearbox designs are evaluated using the surrogate and
analytical models described next.

\subsection{Motor Surrogate and Validation}
Motor electromagnetic performance varies nonlinearly with geometry and winding
because of coupled flux distribution, magnetic saturation, leakage, and loss
mechanisms. Accurate closed-form models over the considered design space are
therefore difficult to obtain, whereas repeated field-based electromagnetic
evaluations would make the system-level search prohibitively expensive. We
consequently use surrogates for motor evaluation.

We construct a separate motor surrogate for each of the seven retained
slot--pole combinations. For each combination, 1000 valid designs are sampled
from the eight-dimensional cross-sectional geometry space using Latin
hypercube sampling (LHS)
\cite{r30,r31}. Motor length and winding choices are applied
subsequently through analytical scaling.

Each surrogate is a standardized quadratic model fitted with ridge regression.
It predicts the electromagnetic and physical quantities required by the
downstream actuator evaluator, including $K_t$, $K_e$, torque, speed,
efficiency, inductance, motor mass, and rotor inertia. Separate models are used
because the slot--pole combinations introduce discrete changes in winding
layout and electromagnetic behavior.

Predictive accuracy is assessed using 200 independent validation cases. For the
representative 18S/20P topology, the $R^2$ values of the eight reported outputs
range from 0.9681 to 0.9993 (Fig.~\ref{l014}), supporting use of the
surrogates for large-scale candidate screening. The selected design is
subsequently re-evaluated before prototype fabrication.

\subsection{Planetary Gearbox Modeling}
Unlike the nonlinear motor electromagnetic response, gearbox kinematics,
efficiency, strength-based sizing, mass, and inertia can be evaluated
efficiently using analytical relations once the topology and tooth-number
combination are specified. We therefore use the analytical model developed
below for gearbox evaluation.

The NGW, NW, and 3K reducers in Fig.~\ref{l012}(b) are distinguished
by their kinematic topology and tooth-number combinations, from which the
reduction ratio is calculated analytically. Here, $z_s$, $z_p$, and $z_r$
denote sun, planet, and ring tooth numbers, respectively; subscripts distinguish
the compound members.

Individual mesh efficiencies $\eta_a$, $\eta_b$, and $\eta_c$ are calculated
from the tooth numbers, mesh type, friction coefficient, and contact geometry
\cite{r13}. With
$q=(z_{p1}/z_s)(z_{r1}/z_{p2})$,
$i_1=z_{r1}/z_s$, and
$i_2=z_{r1}z_{p2}/(z_{r2}z_{p1})$, the forward efficiencies are
\begin{equation}
\begin{aligned}
\eta_{\mathrm{NGW/NW}}
  &=\frac{1+q\eta_a\eta_b}{1+q},\\
\eta_{\mathrm{3K}}
  &=\frac{(1+\eta_a\eta_b i_1)(1-i_2)}
          {(1+i_1)(1-\eta_b\eta_c i_2)}.
\end{aligned}
\label{l015}
\end{equation}
The NGW case follows by setting $z_{p1}=z_{p2}$, while the 3K expression
corresponds to the considered arrangement with $i_2<1$.

For target ratios from 1 to 50 in increments of 0.5, feasible tooth-number
combinations are enumerated and the working center distance, module, and
profile shifts are selected to maximize gearbox efficiency subject to assembly,
clearance, and package constraints. Gear face widths are then sized from the
per-planet load $T_p$ using
\begin{equation}
b_j=
\frac{\sigma_{\mathrm{ref},j}T_p b_{\mathrm{ref}}}
     {T_{\mathrm{ref}}\sigma_{\mathrm{tar}}},
\label{l016}
\end{equation}
where $\sigma_{\mathrm{ref},j}$ is the reference tooth-root stress of mesh pair
$j$ and $\sigma_{\mathrm{tar}}$ is the allowable target stress
\cite{r29}. The sized gears determine reducer mass, axial length,
and inertia. For each topology--ratio pair, the highest-efficiency feasible
candidate is retained for the co-design search. Its actual ratio $i_g$ and
efficiency $\eta_g$ are then used to map the motor torque--speed limits to the
joint output side.
The motor-side rotary inertia is reflected to the joint output by the square of
the reduction ratio,
\begin{equation}
J_{\mathrm{ref,out}}
=i_g^2\left(J_{\mathrm{m,rotor}}+J_{\mathrm{hub}}+J_{\mathrm{sun}}\right),
\label{l017}
\end{equation}
where the three terms are the inertias of the motor rotor, rotor hub, and sun
gear on the gearbox input side, respectively.

\subsection{Actuator Mass Estimation}
Rather than assigning a fixed actuator mass, we recompute it for every candidate
from the motor materials, structural parts, gears, and gearbox structures. Motor
materials follow the candidate geometry and winding, while gear masses follow
the topology-specific geometry and strength-based sizing. Supporting components
are approximated as solid or hollow cylinders and annuli. The same geometric
approximations provide the input-side inertias used in
Eq.~\eqref{l017}. The resulting candidate-specific mass
enters both the torque-density objective and the jumping dynamics.

\subsection{Multi-Objective Co-Design Optimization}
Each enumerated motor--gearbox case is searched using constrained
\mbox{NSGA-II} \cite{r32,r33}. For each decoded hardware candidate, an
inner bounded L-BFGS-B search jointly optimizes 12 B\'ezier coefficients, six
for the hip-torque curve and six for the knee-torque curve.
The resulting policy is evaluated under the candidate-specific torque--speed
limits, contact constraints, and predictive-catch condition in
Eq.~\eqref{l009}. The optimized $h_{\mathrm{jump}}$ and rated
actuator torque density then form the two objectives in
Eq.~\eqref{l013}.

Each enumerated case uses a population of 48 for 30 generations (1488
evaluations including initialization). Standard NSGA-II selection, crossover,
and bounded mutation are applied with feasibility-first ranking; infeasible
candidates are ordered by aggregate constraint violation. Feasible fronts from
all cases are then merged and re-ranked to obtain the global Pareto set.

\FloatBarrier
\section{Optimization Results}
\subsection{Pareto-Front and Design Trade-Off Analysis}

\begin{figure}[!t]
  \centering
  \includegraphics[width=\columnwidth]{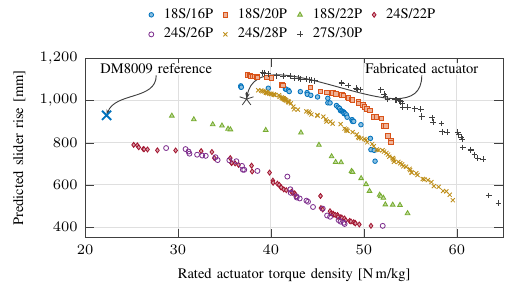}
  \caption{Pareto-front analysis across slot--pole combinations after per-candidate dual-joint B\'ezier-policy optimization. Colored markers denote feasible nondominated candidates. The star and cross show the fabricated actuator and DM8009 reference, respectively.}
  \label{l018}
\end{figure}

Fig.~\ref{l018} shows the feasible trade-off between predicted slider
rise and rated output-torque density across slot--pole combinations when the
hip and knee B\'ezier policies are optimized for every hardware candidate.
Candidates with similar torque density can produce substantially different
task-level performance, indicating that actuator-level metrics alone are
insufficient to rank designs for the jumping task. The fabricated 18S/20P
actuator is shown as a reference rather than as a point on the Pareto fronts.
It was selected from an earlier Pareto front and fabricated before the
co-design workflow was subsequently updated.

\subsection{Actuator- and Task-Level Comparison with the Reference Actuator}
Before fabrication, the selected 18S/20P candidate is locally re-evaluated in
motor simulation software under the final manufacturable winding and
rated-current-density setting. The resulting rated output torque and torque
density are $25.8\zz{N\,m}$ and $37.4\zz{N\,m/kg}$. For convenient
normalization of the hardware commands, the $25.8\zz{N\,m}$ estimate is
rounded to a nominal rated output torque of $26\zz{N\,m}$ in the experiments.
The design is compared with a commercial DM8009 geared joint actuator produced
by DAMIAO Tech., selected as a representative baseline because it is used in
robotic-arm, exoskeleton, and legged-robot platforms, under the same modeled
jumping task and constraints (Table~\ref{l019}) and in the
subsequent hardware validation.

\begin{table}[!t]
\caption{Model Comparison with the DM8009}
\label{l019}
\centering\footnotesize
\begin{tabular*}{\tw}{@{\extracolsep{\fill}}lrr@{}}\toprule
Metric & DM8009 & Optimized\\\midrule
Estimated mass [kg] & 0.896 & 0.689\\
Gear ratio & 9.00 & 32.63\\
Rated output torque [N m] & 20.0 & 25.8\\
Rated torque density [N m/kg] & 22.3 & 37.4\\
Peak torque [N m] & 40.0 & 51.5\\
Maximum speed [rpm] & 320 & 250\\
Predicted slider rise [m] & 0.930 & 1.008\\\bottomrule
\end{tabular*}
\end{table}

The optimized actuator remains lighter and provides higher rated and peak output
torque, yielding an increase of approximately $68\%$ in rated torque density. Its lower maximum
speed still covers the task-relevant operating region. Under the same dynamics
model, the predicted slider rise is $0.930\zz{m}$ for the DM8009 and
$1.008\zz{m}$ for the optimized actuator, an improvement of approximately $8\%$. Thus, additional
high-speed capability does not necessarily improve task performance when the
target motion primarily benefits from greater usable torque.

\section{Experimental Validation}
\subsection{Prototype Design and Implementation}

\begin{figure}[!t]
  \centering
  \includegraphics[width=\columnwidth]{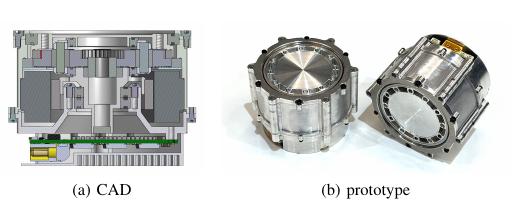}
  \caption{Mechanical design and fabricated prototype of the optimized actuator: (a) CAD sectional view and (b) fully assembled prototype.}
  \label{l020}
\end{figure}

\begin{figure}[!t]
  \centering
  \includegraphics[width=\columnwidth]{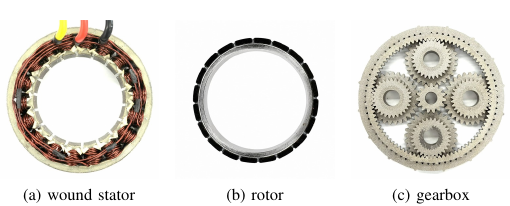}
  \caption{Fabricated key components of the optimized actuator: (a) wound stator, (b) rotor, and (c) planetary gearbox.}
  \label{l021}
\end{figure}

The final motor, planetary gearbox, bearings, housing, and electronics are integrated into a compact coaxial actuator (Fig.~\ref{l020}). The wound stator, rotor, and planetary gearbox confirm that the optimized electromagnetic and transmission designs can be manufactured and assembled as intended (Fig.~\ref{l021}). This implementation converts the optimized design variables into a physical actuator for subsequent experimental validation.

The prototype mass is $0.729\zz{kg}$, $5.8\%$ above the predicted $0.689\zz{kg}$ but $18.6\%$ below the DM8009 mass. At the $26\zz{N\,m}$ rated output-torque setting used in the hardware experiments, its measured-mass torque density is $35.7\zz{N\,m/kg}$. The remaining mass difference reflects practical implementation details not fully represented in the design-stage model.

\subsection{Motor Bench Characterization}

\begin{figure}[!t]
  \centering
  \includegraphics[width=\columnwidth]{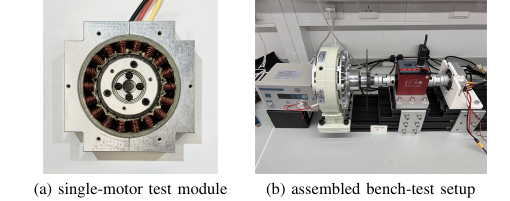}
  \caption{Single-motor bench characterization hardware: (a) fabricated motor test module and (b) assembled setup with the load device and torque sensor.}
  \label{l022}
\end{figure}

\begin{table}[!t]
\caption{Motor Model and Bench Measurements}
\label{l023}
\centering\footnotesize
\begin{tabular*}{\tw}{@{\extracolsep{\fill}}lrr@{}}\toprule
Parameter & Model & Measured\\\midrule
$K_t$ [N m/$A_{\mathrm{line,pk}}$] & 0.04780 & 0.04453\\
$K_e$ [$V_{\mathrm{ll,pk}}$/(rad/s)] & 0.05618 & 0.06391\\
$\omega_0$ [rad/s] & 898.7 & 790\\\bottomrule
\end{tabular*}
\end{table}

The fabricated motor is characterized before gearbox integration using the
single-motor module and bench setup shown in Fig.~\ref{l022}. The tests
compare $K_t$, $K_e$, and no-load speed $\omega_0$ with the model values
(Table~\ref{l023}). These quantities characterize the realized torque
production and speed capability before the complete actuator is evaluated at
the task level. The measured $K_t$ is $6.8\%$ lower, $K_e$ is
$13.8\%$ higher, and no-load speed is $12.1\%$ lower. The fabricated motor thus
retains the principal characteristics predicted by the electromagnetic model,
while measurable deviations remain in the realized hardware. These deviations
are considered when interpreting the model--experiment difference in task performance.

\subsection{Two-DOF Jumping-Leg Experimental Validation}
\begin{figure}[!t]
  \centering
  \includegraphics[width=\columnwidth]{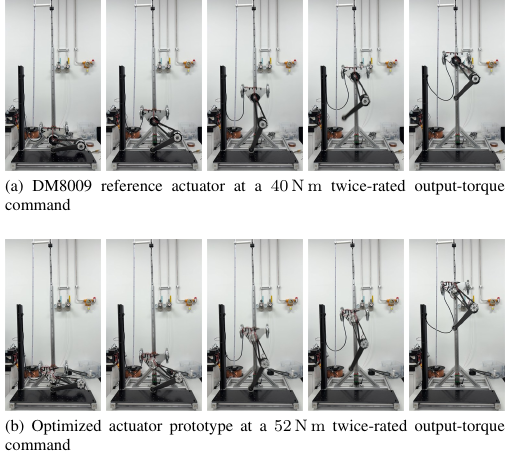}
  \caption{Representative jumping sequences at twice-rated output torque under the common $5.5\zz{kg}$ bench load and $48\zz{V}$ DC bus: (a) DM8009 at $40\zz{N\,m}$ and (b) optimized actuator at $52\zz{N\,m}$.}
  \label{l024}
\end{figure}

The optimized actuator and DM8009 are evaluated on the same vertical-guide
jumping platform under a common $5.5\zz{kg}$ bench load and $48\zz{V}$ DC
bus. Rather than comparing a single torque command, the completed experiment
sweeps the output-torque command relative to each actuator's rated value. The
DM8009 is tested at $15$, $20$, $30$, and $40\zz{N\,m}$, while the optimized
prototype is tested at $19.5$, $26$, $39$, and $52\zz{N\,m}$, corresponding
to $0.75\times$, $1.0\times$, $1.5\times$, and $2.0\times$ rated torque
(Fig.~\ref{l024}).

\begin{figure}[!t]
  \centering
  \includegraphics[width=\columnwidth]{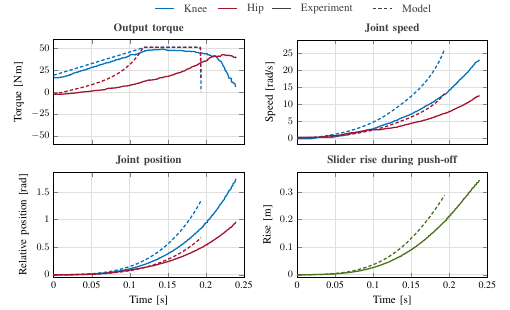}
  \caption{Twice-rated ($52\zz{N\,m}$) push-phase comparison between measured optimized-actuator feedback and the design-stage vertical-guide model prediction. Colors distinguish the knee and hip joints; solid and dashed lines denote experiment and model, respectively.}
  \label{l025}
\end{figure}

\begin{table}[!t]
\caption{Normalized Torque-Sweep Experimental Results}
\label{l026}
\centering\footnotesize
\begin{tabular*}{\tw}{@{\extracolsep{\fill}}lrr@{}}\toprule
Command level & DM8009 & Optimized\\\midrule
$0.75\times$ rated & No jump & 0.490 m\\
$1.00\times$ rated & 0.490 m & 0.580 m\\
$1.50\times$ rated & 0.615 m & 0.745 m\\
$2.00\times$ rated & 0.750 m & 0.840 m\\\bottomrule
\end{tabular*}
\end{table}

\begin{figure}[!t]
  \centering
  \includegraphics[width=\columnwidth]{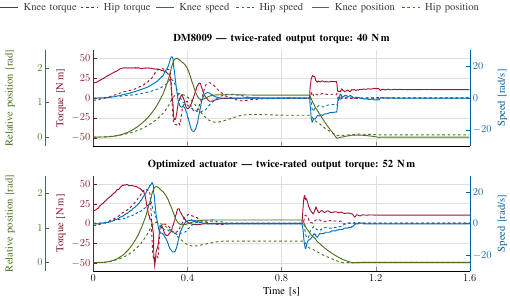}
  \caption{Measured hip and knee responses at twice-rated output-torque commands: $40\zz{N\,m}$ for the DM8009 and $52\zz{N\,m}$ for the optimized actuator. Solid and dashed curves denote knee and hip quantities, respectively.}
  \label{l027}
\end{figure}

Across the normalized sweep, the optimized prototype increases the rise from
$0.490$ to $0.580\zz{m}$ at rated torque and from $0.750$ to $0.840\zz{m}$
at twice-rated torque, corresponding to improvements of $18.4\%$ and $12.0\%$
(Table~\ref{l026}). Figures~\ref{l025} and
\ref{l027} focus on the twice-rated condition. During the push-off phase, the
optimized actuator reaches knee and hip torque peaks of $49.28$ and
$42.96\zz{N\,m}$, respectively, with the larger tracking deficit at the hip.

At this condition, the model predicts an earlier end to the push-off phase than
observed experimentally ($0.194$ versus $0.240\zz{s}$) and less rise during
push-off ($0.287$ versus
$0.343\zz{m}$), yet overpredicts the total rise by $16.6\%$. The discrepancy
is therefore dominated by the post-release prediction. Re-evaluating the same
candidate with the measured motor parameters and actuator mass explains
$11.3$ percentage points, leaving $5.3$ percentage points for unmodeled
transmission, guide, and control effects (Table~\ref{l028}).

\begin{table}[!t]
\caption{Twice-Rated Prediction--Experiment Error Decomposition}
\label{l028}
\centering\footnotesize
\begin{tabular}{lrr}\toprule
Case & Slider rise [m] & Cum. loss [\%]\\\midrule
Design-stage model & 1.008 & 0.0\\
Measured motor parameters & 0.898 & 10.9\\
$+$ measured actuator mass & 0.893 & 11.3\\
Experiment & 0.840 & 16.6\\\bottomrule
\end{tabular}
\end{table}

Additional experimental videos and an interactive illustration of the
co-design framework are available on the project page:
\urlstyle{same}\mbox{\url{https://actuator-codesign.github.io/jump-study/}}.

\newpage
\section{Conclusion and Future Work}
This paper presented a task-oriented motor--gearbox co-design framework for
compact geared actuators. Evidence from representative humanoid motions first
showed that different tasks impose different joint torque--speed requirements.
A tractable two-DOF jumping leg was then used to formulate and validate the
co-design problem. The framework jointly considers motor electromagnetic
design, planetary-gearbox design, actuator mass and inertia, task dynamics, and
control. A hierarchical mixed-variable strategy enables the joint optimization
of discrete and continuous design variables.

The resulting actuator was fabricated and evaluated through motor bench
characterization and jumping experiments against a commercial actuator. The
results show that evaluating coupled motor--gearbox designs through
task-relevant torque--speed demands and task-level performance can produce more
suitable actuators than optimizing isolated component metrics. The comparison
between model predictions and hardware measurements further demonstrates the
importance of propagating manufactured actuator characteristics to task-level
evaluation. Although demonstrated with a planetary-geared actuator, the framework
can be extended to other actuator architectures and to system-level co-design of
robot mechanisms, actuators, and control.

Future work will fabricate and experimentally evaluate a new
actuator selected from the Pareto set and extend validation to coordinated
multi-joint, whole-robot, and multi-task settings.

\FloatBarrier


\begin{thebibliography}{10}
\providecommand{\url}[1]{#1}
\csname url@samestyle\endcsname
\providecommand{\newblock}{\relax}
\providecommand{\bibinfo}[2]{#2}
\providecommand{\BIBentrySTDinterwordspacing}{\spaceskip=0pt\relax}
\providecommand{\BIBentryALTinterwordstretchfactor}{4}
\providecommand{\BIBentryALTinterwordspacing}{\spaceskip=\fontdimen2\font plus
\BIBentryALTinterwordstretchfactor\fontdimen3\font minus
  \fontdimen4\font\relax}
\providecommand{\BIBforeignlanguage}[2]{{%
\expandafter\ifx\csname l@#1\endcsname\relax
\typeout{** WARNING: IEEEtran.bst: No hyphenation pattern has been}%
\typeout{** loaded for the language `#1'. Using the pattern for}%
\typeout{** the default language instead.}%
\else
\language=\csname l@#1\endcsname
\fi
#2}}
\providecommand{\BIBdecl}{\relax}
\BIBdecl

\bibitem{r01}
M.~Hutter, C.~Gehring, D.~Jud, A.~Lauber, C.~D. Bellicoso, V.~Tsounis,
  J.~Hwangbo, K.~Bodie, P.~Fankhauser, M.~Bloesch, R.~Diethelm, S.~Bachmann,
  A.~Melzer, and M.~A. Hoepflinger, ``{ANYmal}: A highly mobile and dynamic
  quadrupedal robot,'' in \emph{Proceedings of the IEEE/RSJ International
  Conference on Intelligent Robots and Systems}, 2016, pp. 38--44.

\bibitem{r02}
Q.~Liao, B.~Zhang, X.~Huang, X.~Huang, Z.~Li, and K.~Sreenath, ``Berkeley
  humanoid: A research platform for learning-based control,'' in
  \emph{Proceedings of the IEEE International Conference on Robotics and
  Automation}, 2025, pp. 2897--2904.

\bibitem{r03}
G.~Ficht and S.~Behnke, ``Bipedal humanoid hardware design: A technology
  review,'' \emph{Current Robotics Reports}, vol.~2, pp. 201--210, 2021.

\bibitem{r04}
K.~Hashimoto, ``Mechanics of humanoid robot,'' \emph{Advanced Robotics},
  vol.~34, no. 21--22, pp. 1390--1397, 2020.

\bibitem{r05}
K.~Hirai, M.~Hirose, Y.~Haikawa, and T.~Takenaka, ``The development of {Honda}
  humanoid robot,'' in \emph{Proceedings of the IEEE International Conference
  on Robotics and Automation}, vol.~2, 1998, pp. 1321--1326.

\bibitem{r06}
M.~Hirose and K.~Ogawa, ``{Honda} humanoid robots development,''
  \emph{Philosophical Transactions of the Royal Society A: Mathematical,
  Physical and Engineering Sciences}, vol. 365, no. 1850, pp. 11--19, 2007.

\bibitem{r07}
H.~Hirukawa, F.~Kanehiro, K.~Kaneko, S.~Kajita, K.~Fujiwara, Y.~Kawai
  \emph{et~al.}, ``Humanoid robotics platforms developed in {HRP},''
  \emph{Robotics and Autonomous Systems}, vol.~48, no.~4, pp. 165--175, 2004.

\bibitem{r08}
Y.~Ogura, H.~Aikawa, K.~Shimomura, H.~Kondo, A.~Morishima, H.~O. Lim, and
  A.~Takanishi, ``Development of a new humanoid robot {WABIAN-2},'' in
  \emph{Proceedings of the IEEE International Conference on Robotics and
  Automation}, 2006, pp. 76--81.

\bibitem{r09}
G.~A. Pratt and M.~M. Williamson, ``Series elastic actuators,'' in
  \emph{Proceedings of the IEEE/RSJ International Conference on Intelligent
  Robots and Systems}, 1995, pp. 399--406.

\bibitem{r10}
S.~Seok, A.~Wang, M.~Y. Chuah, D.~J. Hyun, J.~Lee, D.~M. Otten, J.~H. Lang, and
  S.~Kim, ``Design principles for energy-efficient legged locomotion and
  implementation on the {MIT} cheetah robot,'' \emph{IEEE/ASME Transactions on
  Mechatronics}, vol.~20, no.~3, pp. 1117--1129, 2015.

\bibitem{r11}
P.~M. Wensing, A.~Wang, S.~Seok, D.~M. Otten, J.~H. Lang, and S.~Kim,
  ``Proprioceptive actuator design in the {MIT} cheetah: Impact mitigation and
  high-bandwidth physical interaction for dynamic legged robots,'' \emph{IEEE
  Transactions on Robotics}, vol.~33, no.~3, pp. 509--522, 2017.

\bibitem{r12}
B.~Katz, J.~Di~Carlo, and S.~Kim, ``Mini cheetah: A platform for pushing the
  limits of dynamic quadruped control,'' in \emph{Proceedings of the IEEE
  International Conference on Robotics and Automation}, 2019, pp. 6295--6301.

\bibitem{r13}
H.~Matsuki, K.~Nagano, and Y.~Fujimoto, ``Bilateral drive gear---a highly
  backdrivable reduction gearbox for robotic actuators,'' \emph{IEEE/ASME
  Transactions on Mechatronics}, vol.~24, no.~6, pp. 2661--2673, 2019.

\bibitem{r14}
H.~Zhu, C.~Nesler, N.~Divekar, V.~Peddinti, and R.~D. Gregg, ``Design
  principles for compact, backdrivable actuation in partial-assist powered knee
  orthoses,'' \emph{IEEE/ASME Transactions on Mechatronics}, vol.~26, no.~6,
  pp. 3104--3115, 2021.

\bibitem{r15}
J.~He and F.~Gao, ``Mechanism, actuation, perception, and control of highly
  dynamic multilegged robots: A review,'' \emph{Chinese Journal of Mechanical
  Engineering}, vol.~33, no.~1, p.~79, 2020.

\bibitem{r16}
J.~Di~Carlo, P.~M. Wensing, B.~Katz, G.~Bledt, and S.~Kim, ``Dynamic locomotion
  in the {MIT} cheetah 3 through convex model-predictive control,'' in
  \emph{Proceedings of the IEEE/RSJ International Conference on Intelligent
  Robots and Systems}, 2018, pp. 1--9.

\bibitem{r17}
Y.~Ding and H.-W. Park, ``Design and experimental implementation of a
  quasi-direct-drive leg for optimized jumping,'' in \emph{Proceedings of the
  IEEE/RSJ International Conference on Intelligent Robots and Systems}, 2017,
  pp. 300--305.

\bibitem{r18}
G.~Fadini, S.~Kumar, R.~Kumar, T.~Flayols, A.~Del~Prete, J.~Carpentier, and
  P.~Sou{\`e}res, ``Co-designing versatile quadruped robots for dynamic and
  energy-efficient motions,'' \emph{Robotica}, vol.~42, no.~6, pp. 2004--2025,
  2024.

\bibitem{r19}
Z.~Wu, K.~Zheng, Z.~Ding, and H.~Gao, ``A survey on legged robots: Advances,
  technologies and applications,'' \emph{Engineering Applications of Artificial
  Intelligence}, vol. 138, p. 109418, 2024.

\bibitem{r20}
L.~Zhou, S.~Bai, and M.~R. Hansen, ``Design optimization on the drive train of
  a lightweight robotic arm,'' \emph{Mechatronics}, vol.~21, no.~3, pp.
  560--569, 2011.

\bibitem{r21}
G.~Fadini, T.~Flayols, A.~Del~Prete, N.~Mansard, and P.~Sou{\`e}res,
  ``Computational design of energy-efficient legged robots: Optimizing for size
  and actuators,'' in \emph{Proceedings of the IEEE International Conference on
  Robotics and Automation}, 2021, pp. 9898--9904.

\bibitem{r22}
T.~Dinev, C.~Mastalli, V.~Ivan, S.~Tonneau, and S.~Vijayakumar, ``A versatile
  co-design approach for dynamic legged robots,'' in \emph{Proceedings of the
  IEEE/RSJ International Conference on Intelligent Robots and Systems}, 2022,
  pp. 10\,343--10\,349.

\bibitem{r23}
N.~V. Queipo, R.~T. Haftka, W.~Shyy, T.~Goel, R.~Vaidyanathan, and P.~K.
  Tucker, ``Surrogate-based analysis and optimization,'' \emph{Progress in
  Aerospace Sciences}, vol.~41, no.~1, pp. 1--28, 2005.

\bibitem{r24}
Y.~Jin, ``Surrogate-assisted evolutionary computation: Recent advances and
  future challenges,'' \emph{Swarm and Evolutionary Computation}, vol.~1,
  no.~2, pp. 61--70, 2011.

\bibitem{r25}
E.~Todorov, T.~Erez, and Y.~Tassa, ``{MuJoCo}: A physics engine for model-based
  control,'' in \emph{Proceedings of the IEEE/RSJ International Conference on
  Intelligent Robots and Systems}, 2012, pp. 5026--5033.

\bibitem{r26}
J.~F. Gieras and J.-X. Shen, \emph{Modern Permanent Magnet Electric Machines:
  Theory and Control}.\hskip 1em plus 0.5em minus 0.4em\relax CRC Press, 2023.

\bibitem{r27}
M.~A. Jalali~Kondelaji, M.~R. Sarshar, P.~Asef, and M.~Mirsalim, ``A
  multi-tooth interactive flux reversal permanent magnet motor,'' \emph{IEEE
  Transactions on Industrial Electronics}, 2026, early access.

\bibitem{r28}
J.~Pyrh{\"o}nen, T.~Jokinen, and V.~Hrabovcov{\'a}, \emph{Design of Rotating
  Electrical Machines}, 2nd~ed.\hskip 1em plus 0.5em minus 0.4em\relax Wiley,
  2014.

\bibitem{r29}
{International Organization for Standardization}, \emph{{ISO} 6336-3:2019},
  International Organization for Standardization, 2019.

\bibitem{r30}
M.~D. McKay, R.~J. Beckman, and W.~J. Conover, ``A comparison of three methods
  for selecting values of input variables in the analysis of output from a
  computer code,'' \emph{Technometrics}, vol.~21, no.~2, pp. 239--245, 1979.

\bibitem{r31}
M.~Stein, ``Large sample properties of simulations using latin hypercube
  sampling,'' \emph{Technometrics}, vol.~29, no.~2, pp. 143--151, 1987.

\bibitem{r32}
K.~Deb, A.~Pratap, S.~Agarwal, and T.~Meyarivan, ``A fast and elitist
  multiobjective genetic algorithm: {NSGA-II},'' \emph{IEEE Transactions on
  Evolutionary Computation}, vol.~6, no.~2, pp. 182--197, 2002.

\bibitem{r33}
H.~Zhao, C.~Liu, Z.~Song, and W.~Wang, ``Exact modeling and multiobjective
  optimization of vernier machines,'' \emph{IEEE Transactions on Industrial
  Electronics}, vol.~68, no.~12, pp. 11\,740--11\,751, 2021.

\end{thebibliography}
\end{document}